# Negation Blindness in Large Language Models: Unveiling the 'NO Syndrome' in Image Generation

**Mohammad Nadeem**
Department of Computer Science
Aligarh Muslim University
Aligarh, India
mnadeem.cs@amu.ac.in

**Shahab Saquib Sohail**[*]
School of Computing Science and Engineering
VIT Bhopal University
Sehore, MP, 466114, India
shahabsaquibsohail@vitbhopal.ac.in

**Erik Cambria**
College of Computing and Data Science
Nanyang Technological University, Singapore
cambria@ntu.edu.sg

**Björn W. Schuller**
Technical University of Munich, Germany
Imperial College London, UK
schuller@tum.de

**Amir Hussain**
School of Computing, Engineering and the Built Environment
Edinburgh Napier University, Scotland, UK
a.hussain@napier.ac.uk

September 4, 2024

## Abstract

Foundational Large Language Models (LLMs) have changed the way we perceive technology. They have been shown to excel in tasks ranging from poem writing and coding to essay generation and puzzle solving. With the incorporation of image generation capability, they have become more comprehensive and versatile AI tools. At the same time, researchers are striving to identify the limitations of these tools to improve them further. Currently identified flaws include hallucination, biases, and bypassing restricted commands to generate harmful content. In the present work, we have identified a fundamental limitation related to the image generation ability of LLMs, and termed it "The NO Syndrome". This negation blindness refers to LLMs inability to correctly comprehend 'NO' related natural language prompts to generate the desired images. Interestingly, all tested LLMs including GPT-4, Gemini, and Copilot were found to be suffering from this syndrome. To demonstrate the generalization of this limitation, we carried out simulation experiments and conducted entropy-based and benchmark statistical analysis tests on various LLMs in multiple languages, including English, Hindi, and French. We conclude that the NO syndrome is a significant flaw in current LLMs that needs to be addressed. A related finding of this study showed a consistent discrepancy between image and textual responses as a result of this NO syndrome. We posit that the introduction of a 'negation context-aware' reinforcement learning based feedback loop between the LLM's textual response and generated image could help ensure the generated text is based on both the LLM's correct contextual understanding of the negation query and the generated visual output.

---

[*]Corresponding author

# 1 Introduction

Generative-AI enabled large language models (LLMs), exemplified by OpenAI's GPT-3.5 and GPT-4 [1], Google's Gemini [2], Meta's Llama, etc., have reshaped the landscape of artificial intelligence. These models excel in generating coherent text [3] and have expanded into multimodal domains [4], handling inputs across text [5], images [6], audio [7], and video [8]. Their capabilities also extend to logical reasoning [9] enabling them to effectively tackle a range of tasks, from writing poetry to managing complex generative tasks in different languages [10, 11].

The language models are pivotal in driving technological and societal transformations, raising both economic and ethical considerations across various sectors [12, 13]. The rapid growth in research related to LLMs and generative AI not only highlights the importance of such technologies in the current digital age but also emphasizes the necessity for a comprehensive understanding of their mechanisms, applications, and limitations. The scope of academic and industrial research into LLMs, as highlighted by the substantial volume of scholarly activity (with 11,645 documents related to 'Large Language Models' and 8,643 documents pertaining to 'ChatGPT'), underscores the importance and broad applicability of LLMs, as shown in Figure 1. As the exploration into the vast capabilities and extensive applications of LLMs deepens, it becomes imperative to critically examine and refine them to meet evolving standards and expectations.

Despite their capabilities, LLMs still exhibit limitations, and are susceptible to generating 'hallucinated' content—data that is either fabricated or misleading [14]. The phenomenon poses significant risks, especially when such models are deployed in critical information dissemination contexts and lead to the spread of misinformation [15]. For instance, models like ChatGPT have been observed to fabricate information, creating outputs that might appear plausible yet are fundamentally incorrect. It compromises the generated content's reliability and raises substantial ethical concerns regarding the propagation of biases and stereotypes [16], potentially amplifying harmful narratives [10].

To that end, it is imperative to examine the limitations of LLMs, particularly in terms of accuracy and bias [17]. It helps address the challenge of misinformation by improving the LLMs' ability to detect complex factual inconsistencies to enhance trust in their outputs [18]. Understanding these limitations also helps mitigate the impact of spurious biases to ensure that LLMs are deployed responsibly and effectively in real-world applications [19]. Among many, hallucination and biased responses are the most commonly studied limitations [20]. However, they are qualitative in nature which makes it difficult to assess the degree of their presence. Consequently, researchers are increasingly focusing on developing robust quantitative metrics to effectively measure and analyze these qualitative attributes [14].

For instance, Semantic entropy, an entropy-based uncertainty estimator for LLMs, was devised to detect confabulations (a subset of hallucinations) [15]. It mitigates the variability in expressing a single idea by assessing uncertainty at the semantic level rather than through specific word sequences. The metric demonstrated cross-dataset and cross-task applicability without necessitating prior knowledge of the task, relies on no task-specific data, and exhibits robust generalization to previously unseen tasks. Moreover, a 'moral direction' metric was developed to assess the biases in LLMs using techniques such as Principal Component Analysis (PCA) [16]. The metric enabled the assessment of the normativity (or non-normativity) of arbitrary phrases without the need for explicit language model training for the specific task.

Although the limitations of LLMs are under continuous study, many shortcomings remain unidentified [21]. As progress in the field slowly unveils these limitations, our work contributes by identifying a pervasive flaw in foundational LLMs, which we term the 'NO Syndrome'. The issue arises when LLMs fail to correctly process negations in prompts, a basic yet crucial aspect of natural language understanding. Our investigations reveal that when LLMs are prompted to generate images with specific negations, such as 'a person without spectacles,' they often produce incorrect results, such as images of people with spectacles. The flaw persists across multiple languages, including English, Hindi, and French, suggesting a fundamental limitation in the current models' processing capabilities. By exploring the shortcomings, we aim to enhance the LLMs' factual accuracy and reliability, thereby fortifying trust in AI-generated content and ensuring their more responsible deployment in real-world applications [22].

Furthermore, to enhance the depth and rigor of our analysis, we experimented with various prompts and selected the most promising, common, and straightforward queries. To generalize the obtained results, we tested prompts in three different languages: English, Hindi, and French. Our study highlights a limitation that appears to be present in all foundational LLMs, regardless of the language used. To quantify inaccuracies, we employed two metrics: the percentage of inaccuracy and entropy. The metrics offer a robust means to assess the extent of inaccurate results and compare the performance of LLMs effectively. Additionally, we conducted non-parametric statistical tests to determine if the outcomes of different LLMs differ significantly. Noteworthy observations include discrepancies between image and textual responses, and prompt types that lead to inaccuracies. The comprehensive approach ensures a thorough and reliable analysis of the performance of LLMs across diverse scenarios.

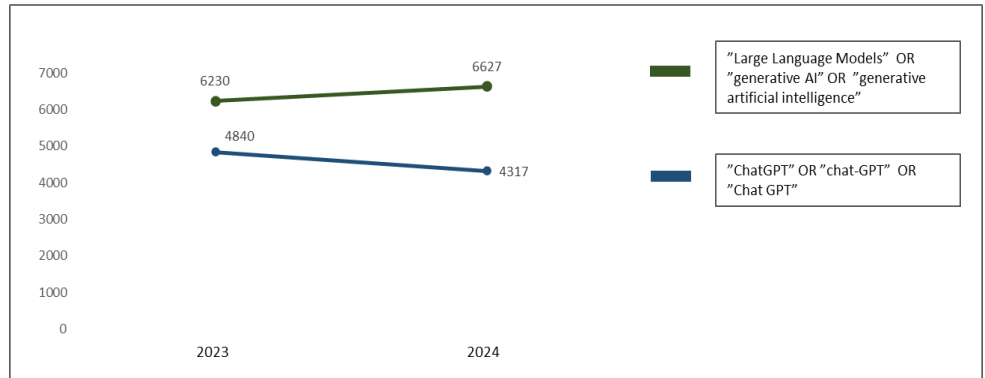

(a)

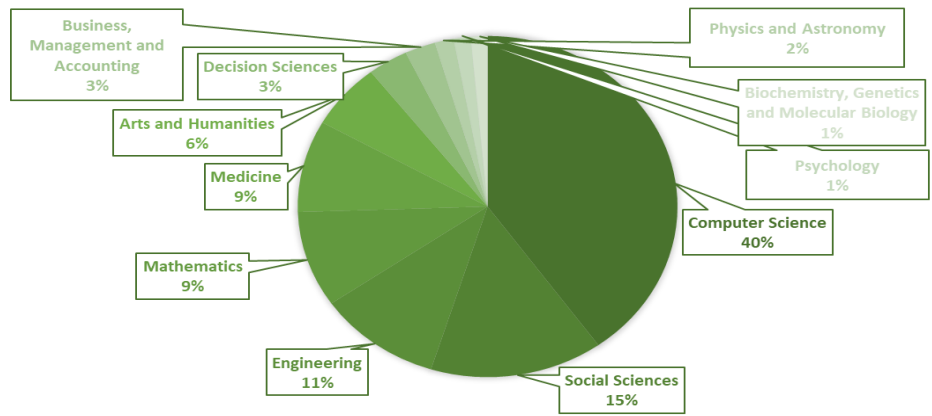

(b)

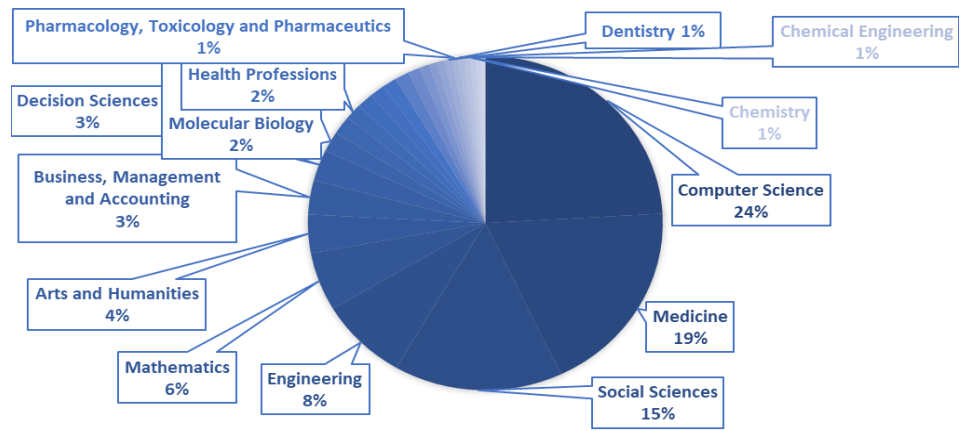

(c)

Figure 1: The growth in research on LLMs. Query on "large language model" OR "generative AI" OR "generative artificial intelligence" for articles published in 2023-24 yields 12,857 results, Whereas query on "ChatGPT" OR "Chat GPT" OR "Chat-GPT" for the same duration appeared to be 9,157, (figure 1a). Additionally, the research has covered many subject areas dominated by Computer Science, Medicine, and humanities, figure 1b and 1c.

## 2 Methodology

In this section, the LLMs employed are discussed. In addition to this, we have provided a detailed account of the adopted methodology (see Figure 2).

### 2.1 LLMs employed

Among many LLMs, we considered GPT-4, Gemini, and Copilot due to their popularity and wide-spread usage. Each of them is developed by leading and reputed organizations (OpenAI, Google, and Microsoft) and ensures the study has a broad and diverse spectrum of LLM technology.

GPT-4 stands for Generative Pre-trained Transformer 4. It is a large language model created by OpenAI, and the fourth in its series of GPT foundation models [1]. Unlike GPT-3.5, GPT-4 is a multimodal model, meaning that it can accept both text and image inputs and can generate both text and image outputs. Gemini is a cutting-edge large language model (LLM) developed by Google. It builds upon Google's T5 transformer model with modifications to enhance its capabilities. Unlike traditional chatbots, Gemini can process and comprehend diverse information formats, including text, images, code, and even audio which allows it to grasp intricate situations with greater nuance and accuracy. Gemini serves as the core technology behind many of Google's AI tools, contributing significantly to their functionality and effectiveness [23].

Copilot is an advanced AI-powered tool designed by Microsoft. Like other LLMs, Copilot can generate text, summarize content, and provide context-specific suggestions within familiar environments such as Word, Excel, PowerPoint, and Outlook. Its sophisticated processing allows it to understand and respond to complex queries, draft documents, create data visualizations, and automate repetitive tasks. At its core, Copilot has been trained on massive datasets of text and code. It can suggest images and generate new images based on the input text description [24].

### 2.2 Languages used

To verify the generalizability of the results, as outlined, we tested the prompts in three languages, namely English, Hindi, and French. The selection of the languages for the experimentation was based on their global significance, broad applicability and linguistic diversity. English is the most widely used language globally and the primary medium for international communication and therefore, provides a broad and accessible base for analysis. Hindi is one of the most used languages in Indian subcontinent and the fourth most spoken language in the world. French is also recognized as a major world language and is used across multiple continents, including Europe, Africa, and North America. These languages have a diverse range of grammatical structures and helped us check the generalizability of the obtained results.

### 2.3 Prompt formation

We designed the following five prompts to get the images generated by LLMs:

- **Q1:** Generate image of a person/dog with no spectacles.
- **Q2:** Generate image of an elephant with no tusks.
- **Q3:** Generate image of flowers with no blue color.
- **Q4:** Generate image of a market with no car.
- **Q5:** Generate image of a car with no dog on top.

The design of the prompts focused on evaluating LLMs capability to handle negative constraints in image generation. Each prompt is crafted to assess the LLM's understanding and execution of exclusion criteria within specific contexts to ensure the model can accurately interpret and generate images and exclude particular elements.

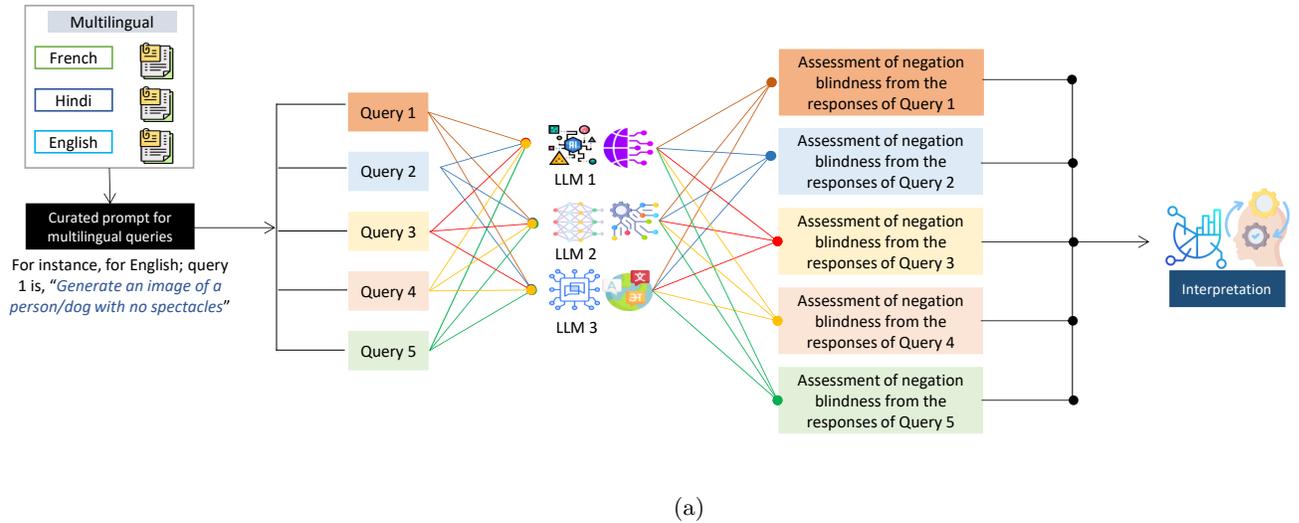

(a)

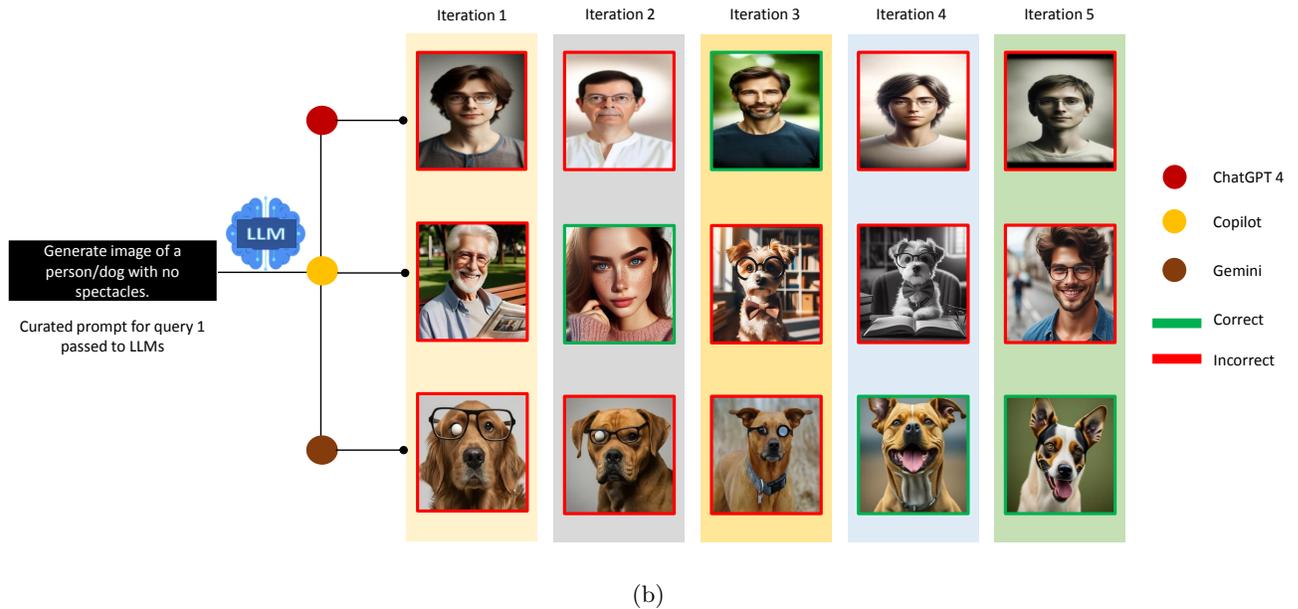

(b)

Figure 2: Figure 2(a) shows the overall methodology adopted in the current work. It outlines the flow of language and query selection, followed by LLMs' responses and interpretation of generated images. Figure 2(b) shows a specific instance of results obtained for English Query-1 (Generate image of a person/dog with no spectacles) with all LLMs. Here, the image with a red border represents an incorrect image, while a green border represents a correct image.

The first prompt tests LLMs ability to identify and exclude a common but optional accessory (spectacles) from people. We included dog also in this query because Gemini refused to produce images of people. The second prompt asks LLMs to exclude an essential body part of the elephant. The third prompt assesses LLMs proficiency in manipulating color properties. The fourth prompt evaluates LLMs for creating complex and dynamic environments while respecting the exclusion of a specific element (cars). The fifth prompt tests the LLMs' precision in interpreting and applying negative constraints in a specific and somewhat uncommon scenario. Overall, all prompts challenged LLMs to adhere to specific exclusion criteria while maintaining the accuracy of the primary concepts.

### 2.4 Response collection

Repeating experiments is a common practice in research to ensure the robustness, reliability, and validity of the findings [25, 26]. Moreover, it improves the performance of ChatGPT [27]. Therefore, we also ran each of five prompts five times in order to ensure the accuracy and reliability of the results. By generating multiple outputs for each prompt, we could assess the consistency of the LLM's performance and identify any potential variations or anomalies in the images produced. It also allowed us to observe whether the model consistently adhered to the exclusion criteria specified in the prompts. Additionally, it helped to ensure that the conclusions drawn from the experiment were based on a comprehensive and representative set of results.

For each run, we stored the image and corresponding textual response generated by LLMs. The overall experimentation process is shown in Algorithm 1.

---

**Algorithm 1** The process of image generation used in the study

---
1: Design relevant prompts with 'NO' keyword.
2: **for** each prompt ($P_i$) **do**
3:    **for** each language ($L_j$) **do**
4:       **for** each Large Language Model (LLM) ($M_k$) **do**
5:          Run prompt $P_i$ in language $L_j$ on LLM $M_k$.
6:          Store the generated image $I_{ijk}$ and record the corresponding textual response ($T_{ijk}$).
7:       **end for**
8:    **end for**
9: **end for**
10: Conduct a performance assessment and compare the results.

---

## 3 Results and Discussion

In this section, we have presented the results and discussed them.

### 3.1 Comparison between LLMs

The results obtained from experiments are presented in Table 1. The results of various Large Language Models (LLMs) demonstrate a clear disparity in their ability to generate accurate images from textual queries. GPT-4 consistently produced a high number of incorrect images across all three languages (English, Hindi, and French), particularly for the queries "Generate an image of flowers with no blue color" and "Generate an image of an elephant with no tusks." It indicates that GPT-4 struggles with understanding and processing negations in the context of image generation. On the other hand, Gemini performs relatively well for English queries but fails to generate any responses for Hindi and French queries. It highlights a significant limitation of Gemini in handling non-English languages in its current version. Copilot exhibited a moderate error in comparison and performed slightly better than GPT-4 on certain queries. However, it also outputs a substantial number of incorrect responses. A sample of incorrect images generated by the LLMs is given in Table 2.

The findings highlight the inherent challenges faced by LLMs in accurately interpreting and generating images from textual descriptions. The high frequency of errors across all LLMs (see Figure 3) suggests that the current state of LLMs is not yet sufficiently advanced to handle complex image generation tasks reliably when dealing with negations or specific exclusions. Moreover, the lack of responses from Gemini for non-English queries underscores the need for LLMs to be more robust and versatile across different languages. In fact, when prompted with Hindi and French queries, it replied with: ``हम इस भाषा में **Gemini** द्वारा चित्र बनाने की

Table 1: Frequency of incorrect images generated by various LLMs.

| Language | Query | LLMs' incorrect responses | | |
|---|---|---|---|---|
| | | GPT-4 | Gemini | Copilot |
| **English** | Generate image of a person/dog with no spectacles. | 4 | 3 | 4 |
| | Generate image of an elephant with no tusks. | 5 | 4 | 5 |
| | Generate image of flowers with no blue color. | 5 | 3 | 4 |
| | Generate image of a market with no car. | 3 | 2 | 3 |
| | Generate image of a car with no dog on top. | 2 | 3 | 4 |
| **Hindi** | ऐसे व्यक्ति/कुत्ते का चित्र बनाएं जिसने चश्मा न पहना हो | 5 | x | 5 |
| | ऐसे हाथी का चित्र बनाएं जिसके दांत न हो | 5 | x | 5 |
| | ऐसे फूलों का चित्र बनाएं जिसमें नीला रंग न हो | 5 | x | 4 |
| | ऐसे बाज़ार का चित्र बनाएं जिसमें कार न हो | 2 | x | 3 |
| | ऐसी कार का चित्र बनाएं जिसके ऊपर कुत्ता न हो | 2 | x | 3 |
| **French** | Générez une image de personne/chien sans lunettes | 2 | x | 5 |
| | Générez l'image d'un éléphant sans défenses | 5 | x | 5 |
| | Générez une image de fleurs sans couleur bleue | 5 | x | 5 |
| | Générer l'image d'un marché sans voiture | 2 | x | 4 |
| | Générez l'image d'une voiture sans chien sur le dessus | 2 | x | 3 |

क्षमता को बेहतर बनाने का प्रयास कर रहे हैं. यह सुविधा शीघ्र ही उपलब्ध कराई जाएगी और रिलीज़ नोट्स को इसके लिए तत्काल अपडेट कर दिया जाएगा" and *"Nous mettons tout en œuvre pour améliorer la capacité de Gemini à générer des images dans cette langue. Cette fonctionnalité devrait bientôt être prête et nous mettrons à jour les notes de version dès qu'elle sera disponible"* respectively. Both the statements loosely translate to: *"We are trying to improve Gemini's ability to draw images in this language. This feature will be made available soon and the release notes will be updated immediately"*.

Overall, while there are variations in performance among LLMs, the results indicate a significant need to enhance accuracy in the domain of image generation.

### 3.2 Entropy calculation

In information theory, Entropy is a measure of the uncertainty or randomness in a set of outcomes [28]. It quantifies the expected amount of information when observing a random variable. For a discrete random variable $X$ with possible outcomes $x_1, x_2, \ldots, x_n$ and a probability mass function $P(X)$, the entropy $H(X)$ is defined as:

$$H(X) = -\sum_{i=1}^{n} P(x_i) log_2 P(x_i). \quad (1)$$

For our work, entropy provides a measure of the uncertainty or variability in the LLMs' responses. Lower entropy values indicate that the LLMs' responses are more predictable and consistent, while higher entropy values suggest greater variability and unpredictability. The entropy values for each LLM are shown in Figure 4.

### 3.3 Statistical Analysis

To test whether the responses of LLMs are significantly different or not, we carried out two statistical tests, namely Friedman Test [29] and Wilcoxon Signed-Rank Test [30]. Both are non-parametric tests [31] and do not assume the data are normally distributed. The Friedman Test is used to detect differences in treatments across multiple test attempts but requires at least three sets of samples. The test calculates the rank of each row (subject) across columns (treatments) and then analyzes them. The value of $p$ associated with the test statistic $F$ obtained is used to verify the hypothesis. Equation (2) presents the formula used to calculate $F$.

$$F = \left[ \frac{12}{b(k)(k+1)} \sum_{l=1}^{k} T_k^2 \right] - 3b(k+1), \quad (2)$$

where, $k$ represents the number of treatments, $b$ is number of rows or blocks, and $T_k^2$ is the squared sum of ranks for sample treatment $k$. The test statistic $F_t$ approximately follows a chi-square distribution with k−1 degrees of freedom.

Table 2: The table shows sample incorrect images generated by various LLMs for different queries in English language.

| Query | Sample images from LLMs | | |
|---|---|---|---|
| | GPT-4 | Gemini | Copilot |
| Q1 | 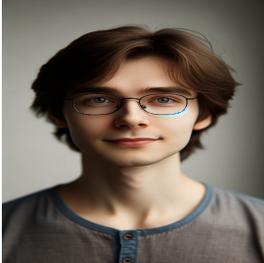 | 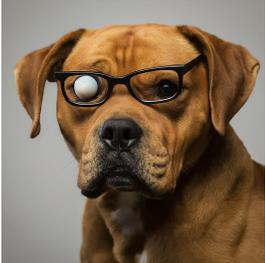 | 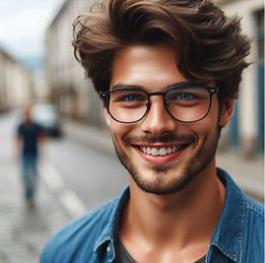 |
| Q2 | 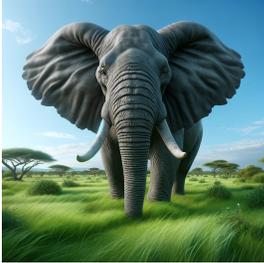 | 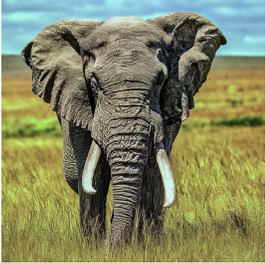 | 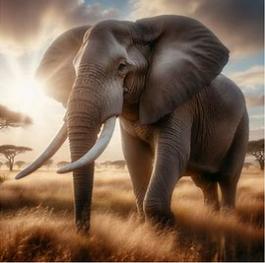 |
| Q3 | 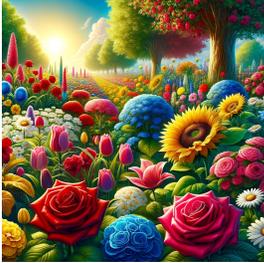 | 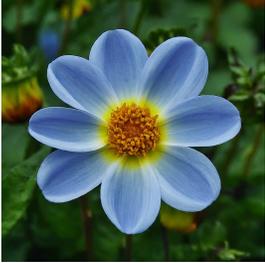 | 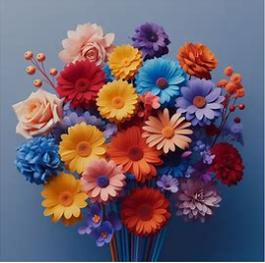 |
| Q4 | 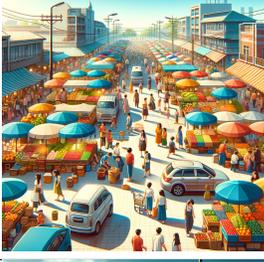 | 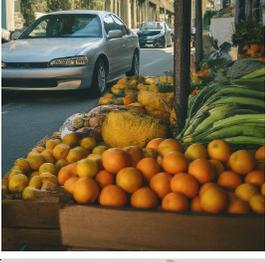 | 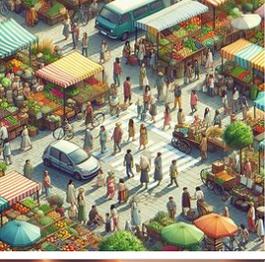 |
| Q5 | 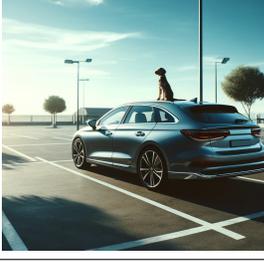 | 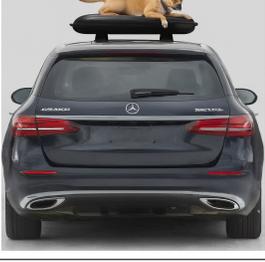 | 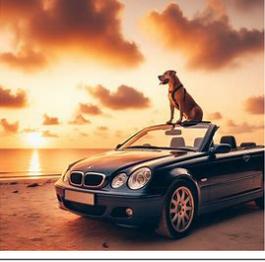 |

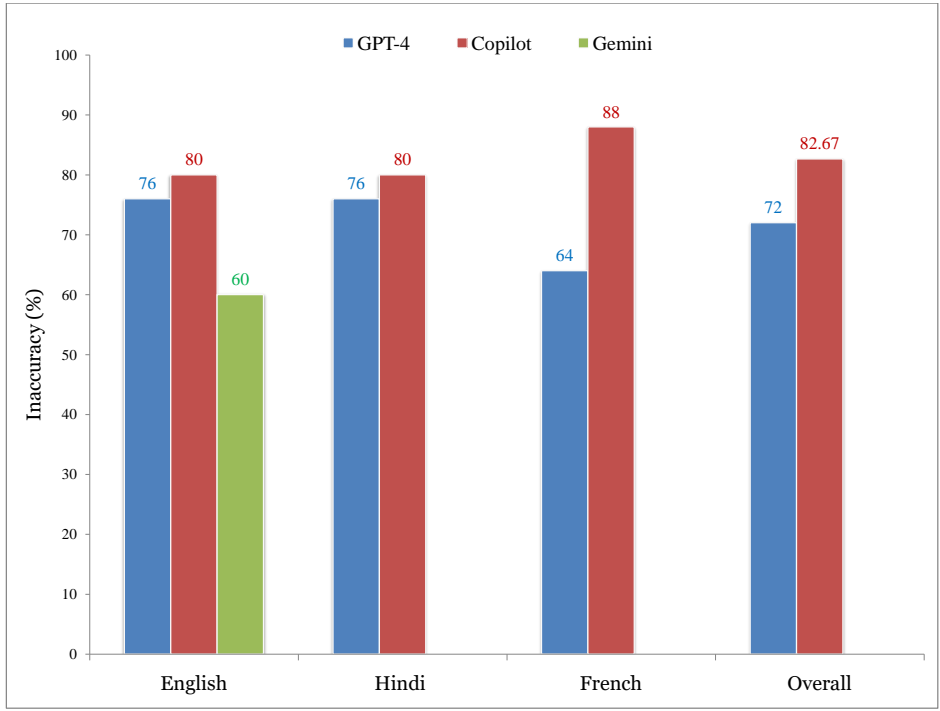

Figure 3: The figure shows the percentage of inaccuracy for different language models (GPT-4, Copilot, Gemini) across different languages (English, Hindi, French) and an overall inaccuracy value. Though Gemini produced the least incorrect results, it could not process Hindi and French languages. On the other hand, Copilot exhibited the highest percentage of inaccuracy. The margin of difference in performance between GPT-4 and Copilot was highest for the French language.

On the other hand, the Wilcoxon signed-rank test can be applied to two related samples. It calculates the differences between paired observations, ranks the differences, sums the ranks of positive differences, and compares the sum to its expected value under the null hypothesis. The test statistic $W$ can be calculated according to Equation (3).

$$W = \sum_{i=1}^{N}[sgn(x_{2,i} - x_{1,i}).R_i], \quad (3)$$

where, $N$ represents the sample size, $sgn$ is the sign function, $x_{2,i}$ and $x_{1,i}$ corresponds to ranked pairs from two distributions, and $R_i$ is the rank of the absolute difference between $x_{2,i}$ and $x_{1,i}$. Moreover, the distribution of $W$ can be approximated by a normal distribution.

Since Gemini produced responses only for English queries, the Friedman Test was applied to the outcomes of all three LLMs for English queries only. The Wilcoxon Signed-Rank Test was applied to GPT-4 and Copilot responses only across all three languages. For the Friedman Test, the test statistic was calculated as 5.76, with 0.056 as the corresponding p-value. The p-value (0.056) is just above the common significance level of 0.05. It suggests a trend towards significant differences in the number of incorrect responses among GPT-4, Gemini, and Copilot for English queries, but it is not statistically significant at the 0.05 level. For the Wilcoxon Signed-Rank Test, the test statistic was 6.0 and the corresponding p-value was 0.084. The p-value (0.084) indicates no statistically significant difference between the number of incorrect GPT-4 and Copilot responses across all languages. However, the p-value is once again relatively close to the 5% significance level to weakly suggest some differences in the responses.

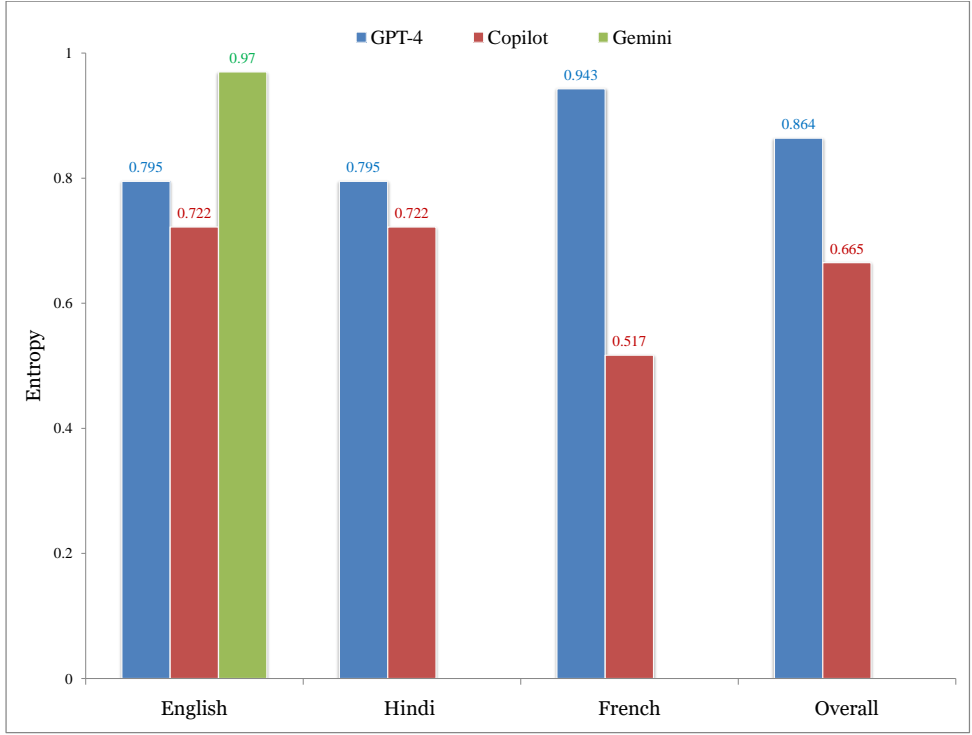

Figure 4: The figure shows entropy values for different language models (GPT-4, Copilot, Gemini) across different languages (English, Hindi, French) and an overall entropy assessment. Copilot provided incorrect responses more consistently in English compared to GPT-4 and Gemini, with Gemini showing a better distribution of correct answers. The trend is similar for Hindi language except Gemini is not yet able to handle queries in Hindi. For French queries, Copilot demonstrates more incorrect responses than GPT-4.

### 3.4 Discrepancy between image and textual response

An interesting observation was that when we asked an LLM to generate an image using keyword 'NO', i.e., not to include a specific element, it often did not generate an accurate image, however, the corresponding textual response indicated that it excluded that element. For example, at a particular instance, when we passed the prompt *Generate image of an elephant with no tusks* to GPT-4, it generated the image shown in Table 2 but the textual response was: *Here is the image of an elephant without tusks, standing in its natural habitat.* There was an obvious misalignment between the text and image generation capabilities of each LLM used. The discrepancy highlights a few cases. Although an LLM's image generation model interprets textual description to create the visual content, text and image generation processes are essentially separate. There is no direct feedback loop between the LLM's textual response and the image generated, i.e., the text is based on the LLM's understanding of the query, not on the final visual output.

It is also important to mention that apart from image generation using the 'No' keyword, we also tried text generation under similar settings but could not find a violation for any LLM. For instance, we supplied the prompt *"Describe 'Earth' with no mention of the word"* multiple times to the LLMs but got the correct response each time. The following was the response of Gemini at a specific instance: *"Imagine a swirling blue marble, suspended in the vast blackness of space. It's mostly covered in vast liquid oceans, teeming with life in all shapes and sizes. White wisps of clouds swirl around its surface, occasionally parting to reveal continents of brown, green, and white – some scorched by deserts, others cloaked in lush forests or vast, icy plains. This sphere spins, bathed in the light of a life-giving star, creating a cycle of light and darkness that nourishes a dizzying array of plants and animals. That's our home, a planet teeming with life and constantly changing."*.

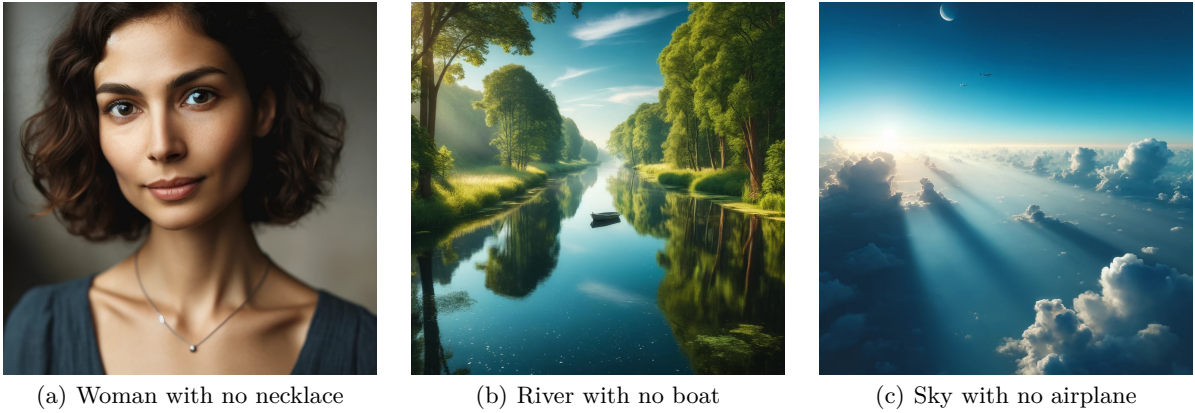

| (a) Woman with no necklace | (b) River with no boat | (c) Sky with no airplane |

Figure 5: The figure shows the images generated by GPT-4 for other similar prompts. It is evident that GPT-4 exhibited the 'NO' syndrome here also and produced incorrect images for similar queries.

### 3.5 Other LLMs and Prompts

To generalize the results further, we tested a few more prompts and one more LLM, i.e., Meta's Llama [32]. The other prompts were the following:

1. Generate image of a woman with no necklace.
2. Generate image of river with no boat.
3. Generate an image of the sky with no airplane.

For these extra prompts also, we got similar violations where the generated images had the components that needed to be excluded. Some sample images of the extra prompts are provided in Figure 5. Meta's Llama-3 is also not free from errors and produces similar results as other LLMs when tested for the original five image generation queries. The images generated through Llama-3 are shown in Figure 6.

### 3.6 Observations

There are a few interesting observations that need further exploration. We already mentioned in Section 4.4 that there is a misalignment between the generated image and the corresponding textual answer. Moreover, it is obvious that changes in language do not have a significant impact on the results produced by any LLM. It is also noteworthy that although we used the 'No' keyword in each prompt, some prompts produced incorrect images more frequently. For the prompt, *Generate an image of an elephant with no tusks*, we never got the accurate image from any LLM in any language. For other prompts such as *Generate an image of a market with no car*, we got fewer inaccurate image instances.

It is also worth mentioning that not every prompt with the 'No' keyword will produce inaccurate results. We tested prompts, such as *Generate image of a mosque from inside with no worshippers*, which almost always produced accurate results (see Figure 7). Therefore, it is difficult to discern beforehand which prompt with exclusion criteria will produce inaccurate results.

### 3.7 Limitations

The study focuses on a few specific LLMs (GPT-4, Gemini, and Copilot). While they are significant models, the findings may not be generalized to all existing or future LLMs. A more comprehensive study that includes other LLMs might increase the acceptability of the results. Moreover, the experiments conducted across multiple languages (English, Hindi, and French) were limited which could limit the reliability of the findings. More languages, queries, and rigorous experimentation would yield better evidence to support the existence of the "NO syndrome".

The study does not provide concrete reasons for the mentioned flaw in image generation and, therefore, could not provide suggestions for its removal. Moreover, LLM capabilities are evolving rapidly with frequent updates and improvements. Upgradation of LLMs after the current study's completion could also render

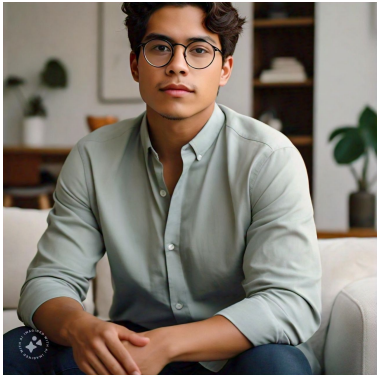
(a) Q1

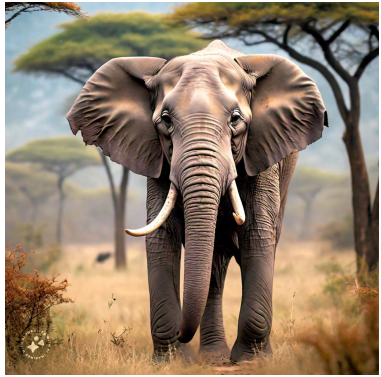
(b) Q2

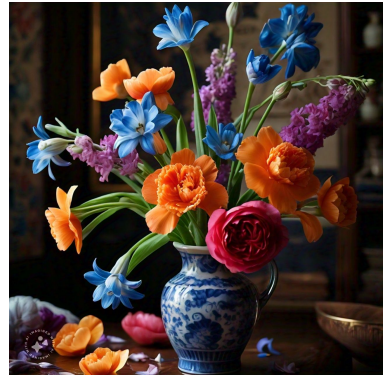
(c) Q3

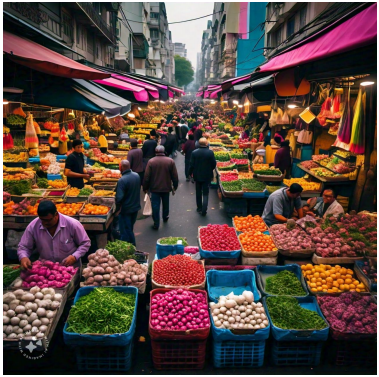
(d) Q4

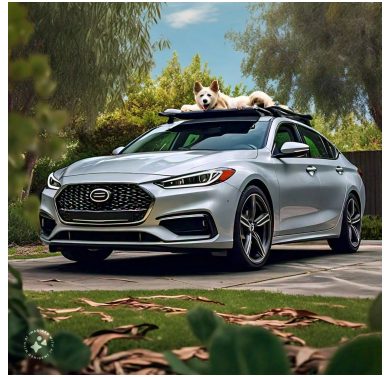
(e) Q5

Figure 6: The images generated by Meta's Llama for the original five prompts show that the performance of Llama follows other LLMs and it also suffers from negation blindness.

some findings less relevant, i.e., findings related to the current limitation may be mitigated in future versions, potentially limiting the long-term applicability of the study's conclusions.

## 4  Conclusion

The present study has explored and highlighted a significant limitation within large language models (LLMs) concerning image generation. It demonstrates LLMs repeated inability to omit specific elements from generated images despite explicit instructions. Our findings have been replicated across multiple state-of-the-art LLMs including GPT-4, Gemini, and Copilot, and across a number of languages, namely English, Hindi, and French. Comparative experiments show that the identified limitation is not isolated to a single model or language but is a universal issue among current foundational LLMs. The study is timely and crucial as it challenges the flexibility and adaptability of LLMs, especially for applications that require precise image generation.

Further, we have identified a related unexplored limitation, specifically, the discrepancy between image and textual responses. We hypothesize this discrepancy may be a result of the "NO syndrome", highlighting fundamental flaws in the current design and training methodologies of LLMs.

Consequently, we conclude these challenges need to be holistically addressed to enhance the utility and effectiveness of multi-lingual and multi-modal LLMs.

Ongoing work aims to address current limitations of our study including the need to scale and systematically test our findings across a wider range of LLMs and languages, and the development of an open standardised multi-lingual evaluation framework. The results of the ongoing study will be made openly available as a benchmark resource for the research community.

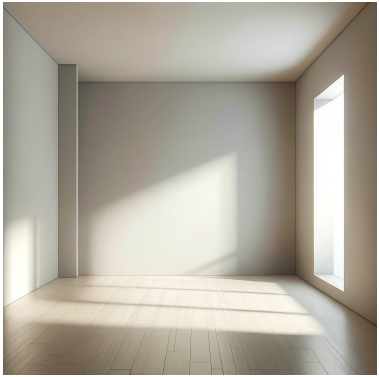 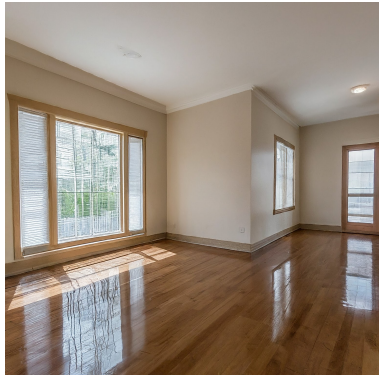 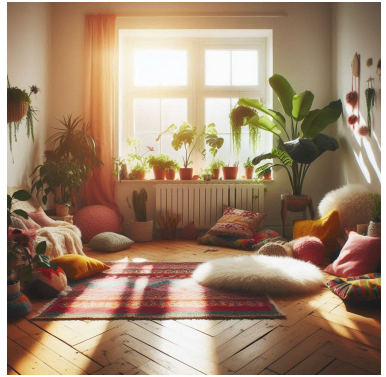

(a) GPT-4: Room with no furniture  (b) Gemini: Room with no furniture  (c) Copilot: Room with no furniture

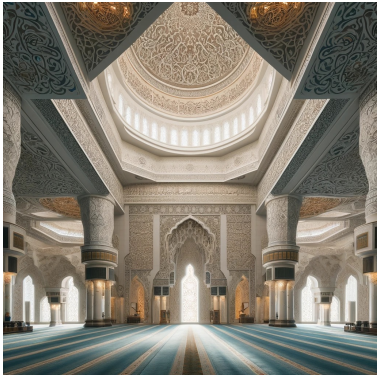 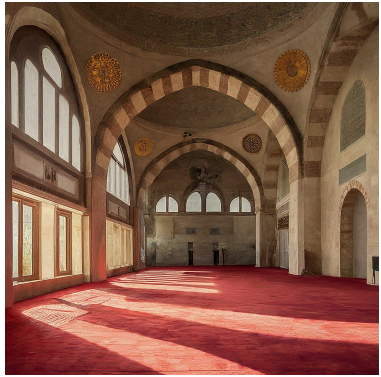 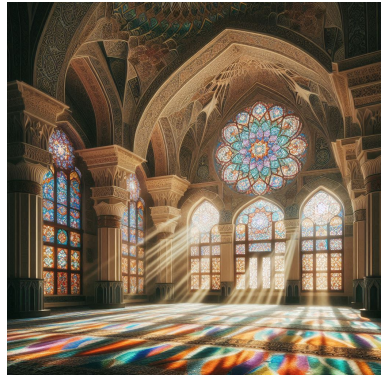

(d) GPT-4: Mosque with no worshippers  (e) Gemini: Mosque with no worshippers  (f) Copilot: Mosque with no worshippers

Figure 7: The figure indicates the results of GPT-4, Gemini, and Copilot for the queries *"Generate image of a room with no furniture"* (a-c) and *"Generate image of a mosque from inside with no worshipers"* (d-f). We tested the prompts multiple times but always got correct images. This indicates that the "No" keyword does not always trick LLMs to generate incorrect results. Therefore, it becomes important to identify the category of queries for which they often generate wrong images.

For future work, we posit that the introduction of a 'negation context-aware' reinforcement learning based feedback loop between the LLM's textual response and generated image could help ensure the generated text is based on both the LLM's correct contextual understanding of the negation query and the generated visual output. The implications of our findings are substantial for developers and researchers in the field of artificial intelligence, implying the need for a greater focus on refining LLMs to handle exclusion criteria more effectively.

## Acknowledgements

Mohammad Nadeem acknowledges the support provided by OpenAI through its Researcher Access Program. Prof Hussain acknowledges the support of the UK Engineering and Physical Sciences Research Council (EPSRC) Grants Ref. EP/T021063/1 (COG-MHEAR) and EP/T024917/1 (NATGEN).